# Can Global Optimization Strategy Outperform Myopic Strategy for Bayesian Parameter Estimation?

Juanping Zhu *[1], Hairong Gu [2]

[1] Maths. Dept., Yunnan University, Kunming, Yunnan, China
[2] Alliance Data, Columbus, OH, USA

**Abstract:**
Bayesian adaptive inference is widely used in psychophysics to estimate psychometric parameters. Most applications used myopic one-step ahead strategy which only optimizes the immediate utility. The widely held expectation is that global optimization strategies that explicitly optimize over some horizon can largely improve the performance of the myopic strategy. With limited studies that compared myopic and global strategies, the expectation was not challenged and researchers are still investing heavily to achieve global optimization. Is that really worthwhile? This paper provides a discouraging answer based on experimental simulations comparing the performance improvement and computation burden between global and myopic strategies in parameter estimation of multiple models. The finding is that the added horizon in global strategies has negligible contributions to the improvement of optimal global utility other than the most immediate next steps (of myopic strategy). Mathematical recursion is derived to prove that the contribution of utility improvement of each added horizon step diminishes fast as that step moves further into the future.

# 1. Introduction

In the context of Bayesian design with nonlinear models and non-Gaussian posteriors, rigorous information-theoretic criteria have been proposed (Lindley, 1956). Bayesian adaptive inference was first applied in the landmark development of the QUEST method by Watson and Pelli (1983). Kontsevich and Tyler (1999) introduced this new Bayesian adaptive method to estimate both threshold and slope of the psychometric function and update posterior probabilities. Bayesian adaptive inference is now widely used in psychophysics to estimate psychometric parameters to explore the connection between physical stimuli and observer's response (Kujala & Lukka，2006, Lesmes, Lu, Baek & Albright, 2010, Lindley, 1956). In addition, Zhu and Zhang (2015) applied Bayesian adaptive inference to estimate the parameters of gap acceptance function in driving behavior field. By taking advantage of Bayesian adaptive inference, Kandasamy et al. (2017) estimated the posterior distribution of the unknown parameters (e.g. cosmological constants) in bounded continuous cosmological parameter space based on Gaussian Process. Besides, Cavagnaro et al. (2010) extended the methodology of parameter estimation to model selection of multiple memory retention models by treating the model index as a parameter for estimation.

Generally speaking, Bayesian adaptive parameter estimation aims to reduce the uncertainty about parameters in a function, by minimizing the expected entropy, or equivalently, maximizing the mutual information (or information gain) between experimental observations and unknown parameters. This strategy is known as greedy or myopic, which only optimizes the utility of the immediate step at each trial. Since global strategy considers forecasted information in the possible scenarios in the future

(Truong, 2014, Jiang, Chai, Gonzalez & Garnett, 2020), it is usually assumed that there can be substantial room for improvement of performance (Wu & Frazier, 2019) by adopting global optimization strategy for sequential decisions. However, computing a globally optimal policy requires solving Bellman equations, which are generally intractable and therefore hindering wider application (Cashore, Kumarga & Frazier, 2016). Most existing work resorts to severe approximations of future information (Bertsekas, 2012, Jiang, Chai, Gonzalez & Garnett, 2020, Powell, 2011, Wu & Frazier, 2019), which can sabotage performance due to accumulated errors (Yue & Kontar, 2020). Most applications for global optimization are usually addressed by specific assumptions such as one-dimensional, looking-ahead over short horizon or simply defined utility function. For example, Wu and Frazier (2019) presented practical two-step ahead approach for Bayesian optimization problem. Garnett et al. (2011, 2012) defined a unique and simple utility instead of mutual information (Kontsevich & Tyler, 1999, Kujala & Lukka, 2006, Lesmes, Lu, Baek & Albright, 2010) and implemented active searching by looking 2 (or 3) step ahead in the simulations..

Given the expectation of the superiority of global optimization, there has not been much research that successfully quantifies such superiority over myopic strategies in a generalizable manner for Bayesian parameter estimation. In addition, there has not been much research that looks at this issue from the perspective of the tradeoff between the performance improvement and the incremental computational and algorithmic burden that global optimization requires. This study aims to strike for these goals to truly bring an ease of mind to researchers who are still wondering whether it is worthwhile to achieve global optimization.

We approached this question by first comparing 2-trial global strategy without any approximation and myopic strategy on three models from different fields. The utilities of two successive future steps are separated to study on their contributions to the optimal global utility based on experiments. To generalize this insight to more than 2-trials, mathematical recursion is derived for the utility analysis of $T$-trial (T$\geq$ 2) global strategy. In order to make the conclusion sound, the algorithm in the experiments did not conduct any approximation which may bring reduced effects on the global expected utility. The experimental simulations and mathematical recursion proof offer a discouraging answer for the superiority of global optimization strategy.

The next section reviews the mathematical foundations of the myopic and global optimization strategies for Bayesian parameter estimation. Readers who are less interested in this part can skip this section.

## 2. Multiple-Trial Global Strategy for Parameter Estimation

### 2.1 Myopic Bayesian inference for parameter estimation

Suppose empirical heterogeneous psychological function $p(y|\theta, d)$ is given by the probability that the participant takes response $y$ given experimental design $d$, where $\theta$ is the parameter vector to be estimated for the individual participant. The basic idea of the myopic strategy is to minimize the uncertainty about $\theta$, which can be formulated as the expected entropy of parameter prior distribution $p_t(\Theta) = [p_t(\theta)]_{\theta \in \Theta}$ for the $t$-th experimental trial (Kontsevich & Tyler, 1999, Kuss, Ja$\ddot{a}$kel & Wichmann, 2005), where $\Theta$ is the space of parameter vector $\theta$. The mutual information $I(\Theta, Y|d)$ after design $d$ can be defined as the utility for $p_t(\Theta)$ describing the uncertainty reduction.

$$u(d|\, p_t(\Theta)) = I(\Theta, Y|d, p_t(\Theta)) = \iint_{\theta \in \Theta, y \in Y} p(\theta, y|d) \log \frac{p(\theta|d)p(y|d)}{p(\theta, y|d)} dy d\theta$$

$$= \iint_{\theta \in \Theta, y \in Y} p(y|\theta, d) p_t(\theta) \log \frac{p(y|\theta, d)}{\sum_{\theta \in \Theta} p(y|\theta, d) p_t(\theta)} dy d\theta,$$

where $Y$ is the space of response $y$. The strategy is to choose design $d$ that minimizes the expected entropy by maximizing the gain of information after completion of the $t$-th trial, i.e., $d_t^* = \underset{d \in D}{\operatorname{argmax}}\, u(d|\, p_t(\Theta))$. In terms of Bayes rule, the parameter prior distribution regarding the participant's response $y_t \in Y$ can be updated as

$$p_{t+1}(\theta|y_t, d_t^*) := \frac{p_t(\theta)p(y_t|\theta, d_t^*)}{\int_{\theta \in \Theta} p_t(\theta)p(y_t|\theta, d_t^*) d\theta}, \qquad (1)$$

for each $\theta \in \Theta$. This strategy considers to only optimize the utility of the immediate step at each trial, i.e., one-step ahead, so it is myopic and greedy.

## 2.2 Global expected utility for global optimization strategy

Without loss of generality, assume that at the onset we will be allowed to look ahead in the future trails, which entails successively calculating the expected utility of next trials. According to dynamic programming, the global expected utility for looking $T$ trials in the future at the beginning of trial $t$ can be written as

$$\max_{\pi \in \Pi} E\{\sum_{k=0}^{T-1} \gamma^t u(d_{t+k}^\Pi | p_{t+k}(\Theta))\}, \qquad (2)$$

where $\gamma \in (0,1]$ is the discount factor and $\Pi$ is the set of possible policies (Bertsekas, 2012). For convenience, we call it $T$-trial global strategy. Each element $\pi \in \Pi$ corresponds to a policy. The notation $d_{t+k}^\pi$ represents the decision rule that determines the design decision $d_{t+k}$ under policy $\pi$ for future $t+k$-th trial, given prior probability distribution $p_{t+k}(\Theta)$. In other words, the objective is to find the best decision rule to solve Eq.(2). By backward dynamic programming, Eq.(2) can be reformulated as Bellman's equation (Bertsekas, 2012, Jiang, Chai, Gonzalez & Garnett, 2020, Powell, 2011) in Eq.(3-4).

For $k = 0,1 \ldots, T-2$,

$$V_{t+k}(p_{t+k}(\Theta)) = \max_{d_{t+k} \in D} \left( u(d_{t+k}|p_{t+k}(\Theta)) + \gamma E\{V_{t+k+1}(p_{t+k+1}(\Theta|p_{t+k}(\Theta), d_{t+k}))\} \right). \qquad (3)$$

$$V_{t+T-1}(p_{t+T-1}(\Theta)) = \max_{d_{t+T-1} \in D} \left( u(d_{t+T-1}|p_{t+T-1}(\Theta)) \right). \qquad (4)$$

$u(d_{t+k}|p_{t+k}(\Theta))$ is the utility for design $d_{t+k}$ given prior $p_{t+k}(\Theta)$ and $V_{t+k}(p_{t+k}(\Theta))$ is the value of prior distribution $p_{t+k}(\Theta)$. $V_{t+k+1}(p_{t+k+1}(\Theta|p_{t+k}(\Theta), d_{t+k}))$ is the value of prior $p_{t+k+1}(\Theta)$ transited from prior $p_{t+k}(\Theta)$ given design $d_{t+k}$ ($k = 0,1 \ldots, T-2$). Let design sequence $\{d_t^*, d_{t+1}^*, \ldots, d_{t+T-1}^*\}$ be the optimal solution of Eq.(3-4). Therefore, the objective of dynamic programming Eq.(2) with Bayesian mutual information as the reward function is essentially to find decision policy to maximize the global expected utility over finite $T$ experimental trials. Thus, $T$-trial global strategy tries to reduce the global uncertainty about the individual's parameters by maximizing the global expected mutual information (information gain) over the finite horizon.

Considering the global expected utility in Eq.(2), we have two alternative approaches to determine the designs. The first approach, called global $T$-step approach, is to assign the global design sequence $\{d_t^*, d_{t+1}^*, \ldots, d_{t+T-1}^*\}$ to the $t$-th trial till $t+T-1$-th trial. After Bayesian prior updating of $p_{t+k}(\Theta|y_{t+k}, d_{t+k}^*)$ as described in Eq.(1) with observation $y_{t+k}$ for $k = 0,1 \ldots, T-1$, the strategy maximizes the global expected utility for the next $T$ trials at the starting point of $t+T$-th trial and assigns the design sequence $\{d_{t+T}^*, d_{t+T+1}^*, \ldots, d_{t+2T-1}^*\}$ to the $t+T$-th trial up to $t+2T-1$-th trial.

The second approach, called $T$-step ahead approach, is to assign design $d_t^* \in \{d_t^*, d_{t+1}^*, \ldots, d_{t+T-1}^*\}$ to the only $t$-th trial. After prior distribution $p_t(\Theta)$ is updated with observation $y_t$, we compute the optimal global expected utility for next $T$ trials to get design sequence $\{d_{t+1}^*, d_{t+2}^*, \ldots, d_{t+T}^*\}$ and pick up $d_{t+1}^* \in \{d_{t+1}^*, d_{t+2}^*, \ldots, d_{t+T}^*\}$ for the $t+1$-th trial.

Especially when $T = 2$, we call this simplest version as 2-trial global strategy. Because $V_{t+1}(p_{t+1}(\Theta)) = \max_{d_{t+1} \in D}\left(u\left(d_{t+1}|p_{t+1}(\Theta)\right)\right)$, so Eq.(3) can be easily rewritten as
$$V_t(p_t(\Theta)) = \max_{d_t \in D}(u(d_t|p_t(\Theta)) + \gamma E\{V_{t+1}(p_{t+1}(\Theta))\})$$
$$= \max_{d_t \in D}\left(u(d_t|p_t(\Theta)) + \gamma E\{\max_{d_{t+1} \in D}(u(d_{t+1}|p_{t+1}(\Theta|d_t, p_t(\Theta))))\}\right)$$
$$= \max_{d_t \in D}\left(u(d_t|p_t(\Theta)) + \gamma \sum_y p(y|p_t(\Theta), d_t) \max_{d_{t+1} \in D}\{u(d_{t+1}|p_{t+1}(\Theta|d_t, y, p_t(\Theta)))\}\right),$$
(5)

where $p(y|p_t(\Theta), d_t)$ is the probability of taking response $y$ given prior $p_t(\Theta)$ and design $d_t$, and posterior $p_{t+1}(\Theta|d_t, y, p_t(\Theta)) = \frac{p_t(\Theta)p(y|\Theta, d_t)}{\int_{\theta \in \Theta} p_t(\theta)p(y|\theta, d_t)d\theta}$ is Bayesian updating of prior $p_t(\Theta)$ for given design $d_t$ and observation $y$. Accordingly, let design sequence $\{d_t^*, d_{t+1}^*\}$ be the optimal solution of 2-trial global strategy for the $t$-th trial and $t+1$-th trial. We therefore have accordingly global 2-step approach and 2-step ahead approach.

## 3. Three Models for Parameter Estimation

Three models from different areas are applied for parameter estimation in the following simulations to compare global and myopic strategies. This section briefly introduces the three models.

**Heterogenous Gap Acceptance Model**

Heterogenous gap acceptance model is used to describe the behavior of a vehicle driver (or a bicycle rider) when deciding whether to merge to the next traffic lane. Empirical Miller's model (Zhu & Zhang, 2015) of gap acceptance function is defined as

$$Pr\left(\text{accept gap } d|\underline{T}_{cr}, \sigma\right) = \Phi\left(\frac{d - \underline{T}_{cr}}{\sigma}\right),$$

where $\Phi(\cdot)$ is the standard cumulative normal probability function to describe the probability that the driver would merge to the next lane given a distance, or gap $d$, from a car driving ahead in that lane. Parameter $\underline{T}_{cr}$ is the driver's critical gap when facing the first gap and $\sigma$ is the disturbance parameter (Zhu & Zhang, 2015). In an experiment, when a driver encounters the situation with a particular $d$, the driver's response is binary, either merge to the next lane or stay in the current lane. With these responses, heterogenous parameters $\underline{T}_{cr}$ and $\sigma$ can be estimated for each driver.

**Visual Psychometric Model**

Most experimental psychophysics deals with measuring sensitivity thresholds that specify the intensity at which the stimulus is detectable (Kontsevich & Tyler, 1999). One example is contrast sensitivity function that models the probability of detecting a given luminance contrast. The function

$$Pr(d|s,b) = \Psi\left(10^{10^{s}(d-b)}\right) = \frac{1}{\sqrt{2\pi}} \int_{-\infty}^{10^{10^{s}(d-b)}} e^{-\frac{b^2}{2}} dt$$

describes the probability of correct detection of a stimulus intensity $d$, where the intensity $d$ is the controllable experimental design and slope $s$ and threshold $b$ are two parameters to be estimated. The psychometric function $\Psi(\cdot)$ is a cumulative function for Gaussian distribution. At each experimental trial, the binary response of whether a given intensity $d$ is detected is collected.

**Memory Retention Model**

The rate of forgetting over time is one of central issues in memory research. Many different models have been developed to describe the probability of correct recall of words given a certain lag time (Cavagnaro, Myung, Pitt & Kujala, 2010). Exponential memory retention function is one model, defined as

$$Pr(d|a,b) = ae^{-bd},$$

where $a$ and $b$ are model parameters to be estimated and $d$ is the stimulus, lag time. In a typical experiment, each trial consists of 'study phase,' in which a participant is given a list of words to memorize, followed by a 'test phase,' in which retention is assessed by testing how many words the participant can correctly recall from the study list (Cavagnaro, Myung, Pitt & Kujala, 2010). The lag time is the length of time between the study phase and the test phase, which can be controlled by the experimenter.

# 4. Simulation Exploration for Potential Gain from 2-trial Global Strategy

This section applies two approaches of global strategy and one myopic approach to perform parameter estimation for the three models introduced in Section 3. As mentioned in Section 2, two approaches for global strategy are global 2-step approach and 2-step ahead approach, and the myopic approach is one-step ahead approach. The objectives of both global approaches are computed based on 2-trial global strategy, but the designs are assigned in different way, which may generate different performance results in the simulations. No approximation of the parameter distribution space is involved for the global strategies so that the performance of global strategies is not compromised.

Simulations are run on the laptop with configuration in Tab.1 and GPU is used to speed up the computation of matrix. The algorithms are conducted on discretized grids with discount factor $\gamma = 1$. All parameter ranges and grid settings are selected from references (Cavagnaro, Myung, Pitt & Kujala, 2010, Kontsevich & Tyler, 1999, Zhu & Zhang, 2015) as shown in Tab.2.

Table 1 The configuration of the laptop

| CPU | RAM | GPU | MATLAB |
|---|---|---|---|
| i7-8750 | 16GB DDR4 | GeForce GTX1060 6G | 2017a |

Table 2 The settings of the grids

| Model | Parameter 1 | | Parameter 2 | | Design | |
|---|---|---|---|---|---|---|
| | Range | Grids | Range | Grids | Range | Grids |
| Gap acceptance | [5,10] | 20 | [1,5] | 20 | [4,12] | 25 |
| Visual psychometric | [0.7 ,7 ] | 50 | [0,10] | 50 | [0,3] | 50 |
| Memory retention | [0,1] | 20 | [0,1] | 20 | [0,50] | 50 |

We use non-informative uniform prior while initializing an experiment, with the acknowledgement that the choice of priors can usefully influence the testing strategy (Cavagnaro, Myung, Pitt & Kujala, 2010, Kontsevich & Tyler, 1999, Zhu & Zhang, 2015). For each specific true parameter vector $\theta_{true}$, the experiment is replicated to take the average $\hat{\theta}$ for the posterior probability $p_{t+1}(\Theta)$ as the estimated parameter values. To compare the performance of different approaches, the mean squared error (MSE) of the estimated parameters is used as one criterion, which can be quantified by $\text{MSE}(\hat{\theta}) = E[(\hat{\theta} - \theta_{true})^2]$. The second criterion for information gain is defined as the entropy difference between initial prior and posterior from each trial, since maximizing global expected mutual information corresponds to maximizing global expected information gain.

## 4.1 Performance Investigation of Two Strategies

For each model, simulations are run with different sets of true parameter values and with different grid settings for parameters and designs to avoid idiosyncratic results. In what follows, we only present the results using one particular set of true parameter values and only one grid setting for each model. However, the same pattern is observed for different parameter values and different grid settings.

In the simulations, the true driver's gap acceptance function is $\Phi\left(\frac{d-7}{2.004}\right)$ with 10,000 replications. The experiment runs 150 trials.

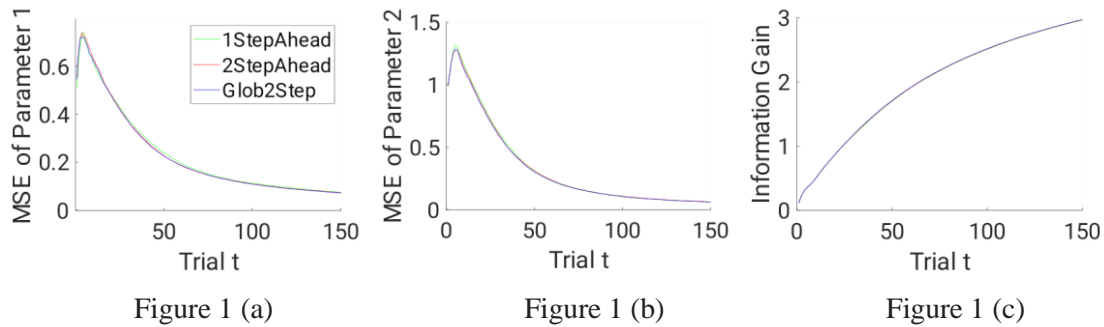

Figure 1 (a)  Figure 1 (b)  Figure 1 (c)

Figure 1 Performance comparison of three approaches for the parameter estimation of the gap acceptance model. (a) is the MSE of parameter $T_{cr}$ and (b) is the MSE of parameter $\sigma$. (c) depicts the information gain with trials.

The true parameter values of visual psychometric probability function are $s = 0.6312$ and $b = 2.3643$ with 5,000 replications and 200 trials for the simulation.

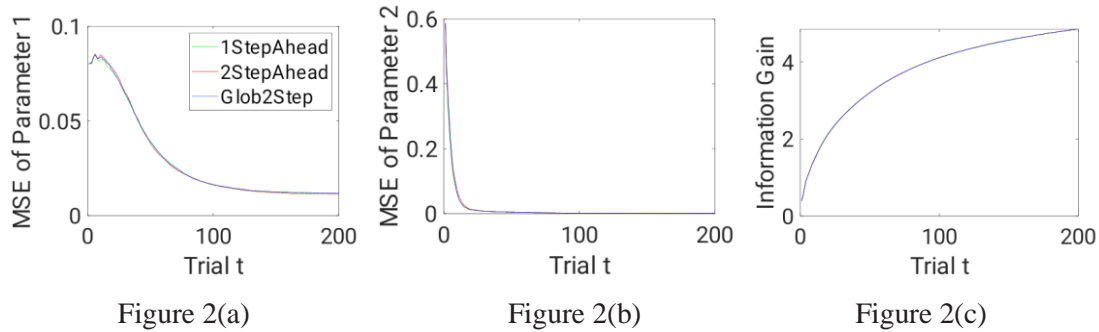

Figure 2(a)　　　　　　　　Figure 2(b)　　　　　　　　Figure 2(c)

Figure 2 Performance comparison of three approaches for the parameter estimation of the visual psychometric model. (a) is the MSE of parameter $s$ and (b) is the MSE of parameter $b$. (c) depicts the information gain with trials.

The true participant Bernoulli memory retention model is selected as $0.7103e^{-.0833t}$ and the participant is given 15 words to memorize. The experiment is replicated 5,000 times and runs 80 trials.

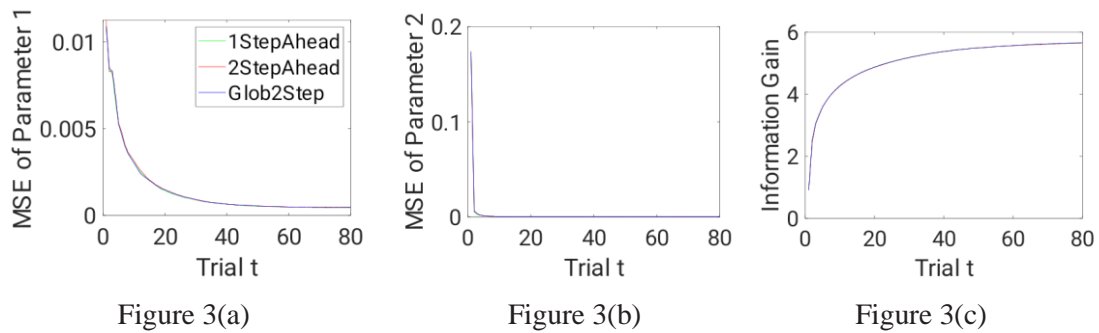

Figure 3(a)　　　　　　　　Figure 3(b)　　　　　　　　Figure 3(c)

Figure 3 Performance comparison of three approaches for the parameter estimation of the memory retention model. (a) is the MSE of parameter $a$ and (b) is the MSE of parameter $b$. (c) depicts the information gain with trials.

Fig.1 regarding gap acceptance mode turn out that MSE of global 2-step approach outperforms 2-step ahead approach slightly, MSE of 2-step ahead approach outperforms myopic approach slightly and three information gain curves almost overlap. Fig.2 for visual psychometric model and Fig.3 for memory retention model display that three MSE curves for three approaches intersect with each other and three information gain curves overlap too.

Overlapping of information gain curves of three approaches in three cases turns out that no approach performs apparently better than another. MSE curves in Fig.1-3 demonstrate that the superiority of 2-trial global strategy is so marginal over myopic strategy that we can even ignore it. Simple 2-trial global strategy implemented for three cases from different fields couldn't apparently outperform myopic approach under rigorous mathematical inference. These results trigger us to deeply investigate how the mechanism of global optimization strategy works on the expected utilities by the backward computation of dynamic programming.

In order to explore the intrinsic source behind mathematical formulas, we investigate how the utilities are transferred between two successive trials in terms of 2-trial global strategy and figure out which trial plays the most important role in the computation of optimal global expected utility in the simulations. For each specific true parameter setting, the experiment for utility investigation is replicated 500 times to take average for the randomness of observation. The utility ranges for two successive trials as well as difference of utilities between two strategies are studied, according to the backward dynamic programming in Eq.(3-4).

## 4.2 Utility Gain from 2-trial Global Strategy

In order to better understand how the global strategy works on the utility, we consider the contribution to global utility of the $t$-th trial and $t+1$-th trial separately. The utility $u(p_t(\Theta))$ for the $t$-th trial and the expected utility $E\{V_{t+1}(p_{t+1}(\Theta))\}$ for the $t+1$-th trial are calculated for each design $d_t \in D$ according to Eq.(5). The maximal global expected utility of 2-trial global strategy can be computed by enumerating $u(p_t(\Theta)) + E\{V_{t+1}(p_{t+1}(\Theta))\}$ over $d_t \in D$. Let $\bar{u}(d_t|p_t(\Theta))$ be the average of $u(p_t(\Theta))$ and $\bar{E}\{V_{t+1}(p_{t+1}(\Theta))\}$ be the average of $E\{V_{t+1}(p_{t+1}(\Theta))\}$ over multiple replications.

Two statistics are calculated to measure how much utility improvement 2-trial global strategy can achieve over myopic strategy. One statistic, utility difference $UD_t$, shows the absolute difference of utilities resulted from two strategies where

$$UD_t = (u(d_t^*|p_t(\Theta)) + E\{u(d_{t+1}^*|p_{t+1}(\Theta|d_t^*, p_t(\Theta)))\}) - (\max_{d \in D} u(d|p_t(\Theta)) + \max_{d \in D} u(d|p_{t+1}(\Theta))), \tag{6}$$

In Eq.(6), term $u(d_t^*|p_t(\Theta)) + E\{u(d_{t+1}^*|p_{t+1}(\Theta|d_t^*, p_t(\Theta)))\}$ is the maximal global expected utility generated from 2-trial global strategy, and term $\max_{d \in D} u(d|p_t(\Theta)) + \max_{d \in D} u(d|p_{t+1}(\Theta))$ is the sum of utilities generated from myopic strategy for the successive $t$-th trial and $t+1$-th trial.

The other statistic, utility difference ratio, is to calculate the relative difference of utilities between two strategies,

$$RD_t = \frac{UD_t}{\max_{d \in D} u(d|p_t(\Theta))}, \tag{7}$$

as the ratio of absolute utility difference $UD_t$ relative to $\max_{d \in D} u(d|p_t(\Theta))$, where $\max_{d \in D} u(d|p_t(\Theta))$ is the maximal utility of the $t$-th trial from myopic strategy. This utility difference ratio can assess how much percentage of myopic $\max_{d \in D} u(d|p_t(\Theta))$ the utility improvement $UD_t$ would take, which can display clearly which trial will decide the value of optimal global expected utility.

**Utility Investigation for Heterogenous Gap Acceptance Function**

For the true driver's function $\Phi\left(\frac{d-7}{2.004}\right)$, Fig.4 presents trials $t = 5, 25, 55, 85, 115, 145$ for instance to display the average utility for the $t$-th trial (in blue) and average expected utility curve for the $t+1$-th trial (in red). For example, the first figure of Fig.4 considers at the starting point of 5-th trial about the average utility for the 5-th trial and the average expected utility for the 6-th trial over design space. The remaining figures in Fig.4 and figures in Fig.6 and Fig. 8 can be explained in the same way.

In Fig.4, the average expected utility curve for the $t+1$-th trial (in red) is much flatter than the average utility curve for the $t$-th trial (in blue) for all $t$. Compared with blue curves, red curves tend more and more flat while the trials move on with $t$. In Fig.5(a), the range width of red curves (of Fig.4) is always far smaller than the range of width of blue curves (of Fig.4) over trial $t$ and the range width of average expected utility for $t+1$-th trial converges to 0. According to Eq.(6), Fig.5(b) shows that the maximal utility difference is $4*10^{-5}$ and the utility difference curve converges to zero. In Fig.5(c), the maximal utility difference ratio is around $1.05*10^{-3}$ and the ratio curve drops dramatically after around 25-th trial and converges to zero, according to Eq.(7).

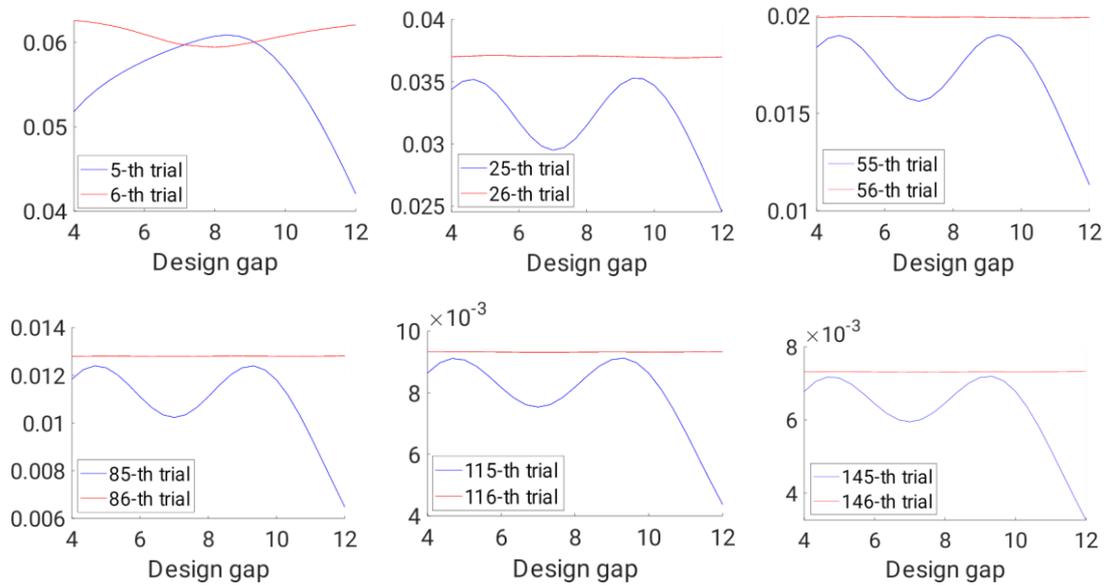

Figure 4 Utility investigation for gap acceptance model at the beginning of each $t$-th trial ($t = 5,25,55,85,115,145$). The blue curve is the average utility for $t$-th trial and the red curve is the average expected utility for $t+1$-th trial.

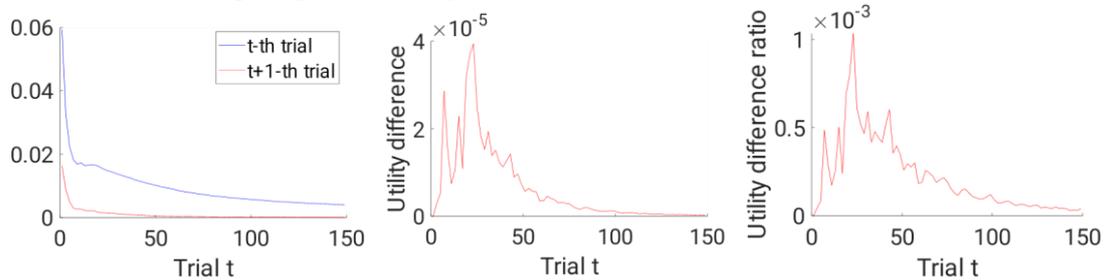

Figure 5(a)           Figure 5 (b)           Figure 5 (c)

Figure 5 (a) The range widths of average utility for $t$-th trial (in blue) and average expected utility for $t+1$-th trial (in red). (b) The utility difference between two strategies. (c) The ratio of utility difference to the maximal $t$-th utility of myopic strategy.

**Utility Investigation of Visual Psychometric Function**

For the true model with $s = 0.6312$ and $b = 2.3643$, Fig.6 presents trials $t = 5, 25, 55, 95, 125, 165$. Similarly, the red curves are much flatter than the blue curves for all $t$-th trial in Fig.6 and the range of average $t$-th trial utility is far wider than the average $t+1$-th trial expected utility as shown in Fig.7(a). In Fig.7(b), the maximal utility difference between two strategies is $8.5*10^{-4}$ and the utility difference curve drops quickly after around the 5-th trial and converges to zero. In Fig.7(c), the maximal ratio of utility difference to the maximal myopic utility is around $5.5*10^{-3}$. And the ratio curve drops dramatically after around 5-th trial and keeps stable around $2.5*10^{-3}$ after the 100-th trial.

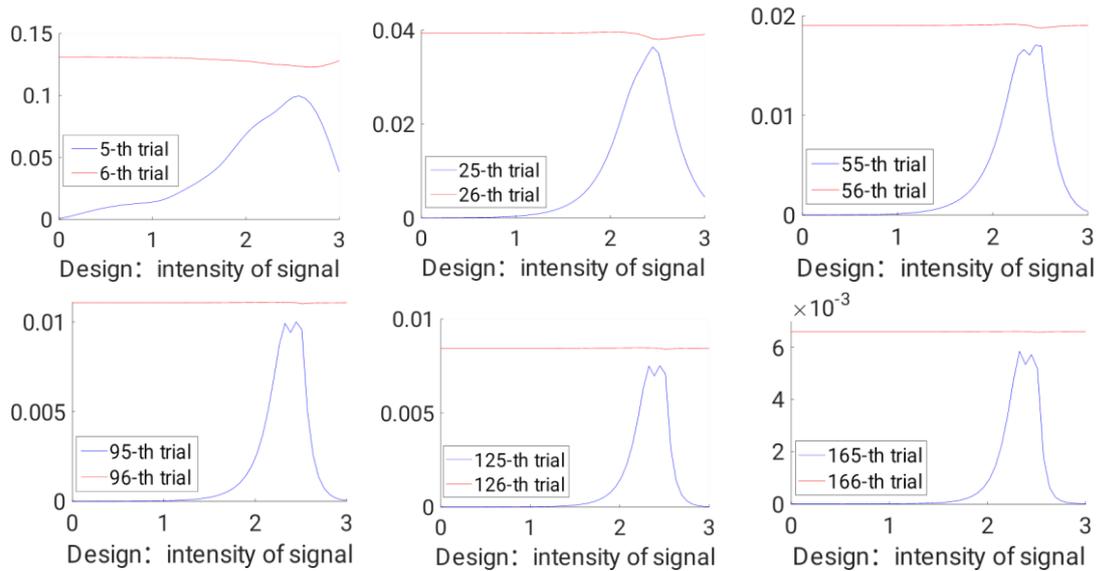

Figure 6 Utility investigation for visual psychometric model at the beginning of each $t$-th trial ($t = 5, 25, 55, 95, 125, 165$). The blue curve is the average utility for $t$-th trial and the red curve is the average expected utility for $t+1$-th trial.

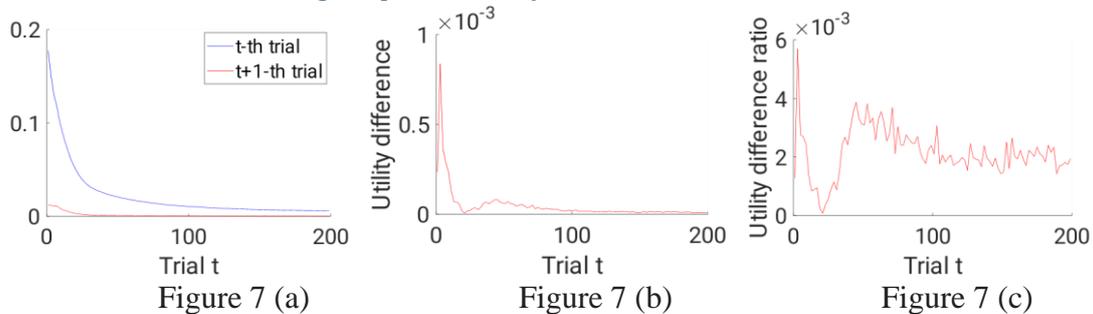

Figure 7 (a) The range widths of average utility for $t$-th trial (in blue) and average expected utility for $t+1$-th trial (in red). (b) The utility difference between two strategies. (c) The ratio of utility difference to the maximal $t$-th utility of myopic strategy.

**Utility Investigation of Memory Retention Function**

Fig.8 presents trials $t = 5, 15, 25, 35, 45, 65$ for true memory retention model $0.7103e^{-.0833t}$. Similarly, the red curves are much flatter than the blue curves for all trials in Fig.8 and the range of average $t$-th trial utility is wider than the average $t+1$-th expected utility as shown in Fig.9(a). In Fig.9(b), the maximal utility difference between two strategies is less than $6.5*10^{-3}$ and the utility difference drops quickly after around the 5-th trial and converges to zero. As shown in Fig.9 (c), the maximal utility difference ratio is around 0.014, and then the curve drops after around the 17-th trial.

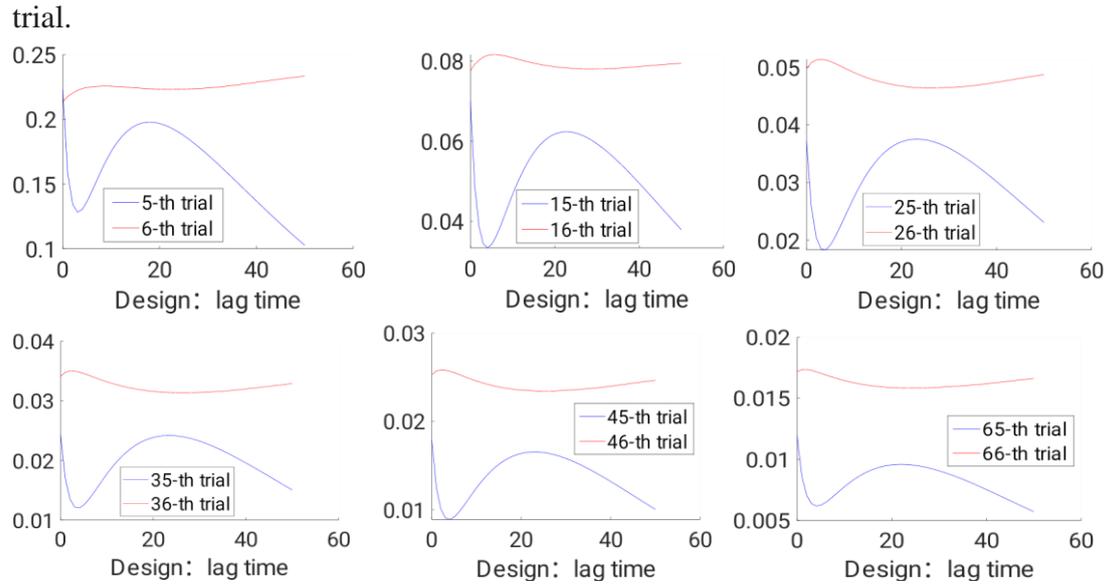

Figure 8 Utility investigation for memory retention model at the beginning of each $t$-th trial ($t = 5, 15, 25, 35, 45, 65$). The blue curve is the average utility for $t$-th trial and the red curve is the average expected utility for $t+1$-th trial.

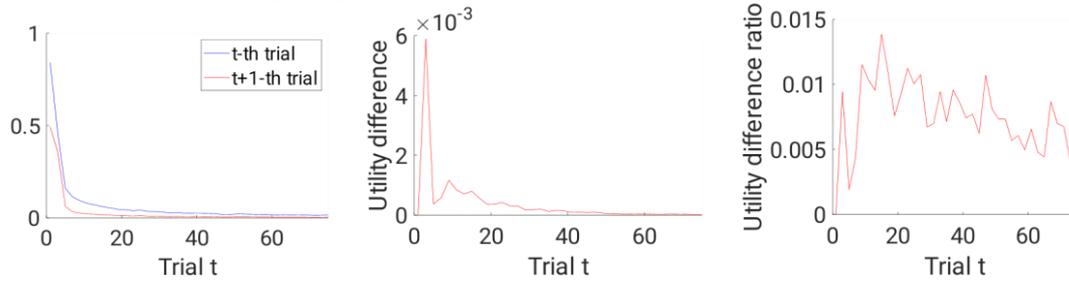

Figure 9 (a)　　　　　　　　Figure 9 (b)　　　　　　　　Figure 9 (c)

Figure 9 (a) The range widths of average utility for $t$-th trial (in blue) and average expected utility for $t+1$-th trial (in red). (b) The utility difference between two strategies. (c) The ratio of utility difference to the maximal $t$-th utility of myopic strategy.

Based on Fig. 4-9, we can summarize the utility investigations from the above three models as follows.

1) The contribution space of $t+1$-th expected utility to the optimal global expected utility $V_t(p_t(\Theta))$ is so narrow that the maximal $t$-th average utility from myopic strategy almost decides the value of optimal $V_t(p_t(\Theta))$ for all trials, according to Eq.(3). This is because the red curve is much flatter than the blue curve, *i.e.*, the range of $t+1$-th average expected utility curve is much narrower than the range of the $t$-th average utility curve in Fig.4,6,8. With the trials moving on, the red curve is getting even flatter and the contribution space is even narrower. The underlying theoretic cause is that the $t+1$-th expected utility is computed on the term $\max_{d_{t+1} \in D} \{u(d_{t+1}|p_{t+1}(\Theta|d_t, y))\}$ in Eq.(5). It is the maximization operation on the $t+1$-th trial that results in the narrow range of $t+1$-th expected utility.

2) The utility difference between two strategies is rather minor so that the performance

improvement from myopic strategy is marginal. Fig.5,7,9(b) show the scales of the utility difference are $10^{-4}, 10^{-5}, 10^{-3}$ and the curve of improvement converges to zero after the narrow peak. This can explain why the information gain curves from three approaches almost overlap as shown in Fig.1-3(c), such that we can even ignore the information gain difference between two strategies.

3) The utility difference ratio is so minor, which indicates the utility difference $UD_t$ between two strategies is negligible compared with value of the maximal $t$-th utility of myopic strategy. As shown in Fig.5,7,9(c), the scales of the ratio $RD_t$ of utility difference to the myopic $t$-th maximal utility are $10^{-3}, 10^{-3}, 10^{-2}$ and the room for performance improvement of 2-trial global strategy over myopic strategy is at most 1.4% of $\max_{d \in D} u(d|p_t(\Theta))$ (of myopic strategy) only for memory retention model but less than 0.6% for the remaining two models. The experimental data deviate far from our assumption that there is substantial room for performance improvement by adopting global optimization strategy for parameter estimation. This can also explain why there is no apparent performance improvement in MSE curves and information gain curves in Fig.1-3 if we use the maximization of the global expected utility (mutual information) as the objective of 2-trial global strategy.

4) Furthermore, computing a globally optimal policy requires solving Bellman equations, which are generally intractable. In our simulations, we sample over all possible $y \in Y$ and $d \in D$ to solve the Bellman equations to get the optimal solution for 2-trial global strategy, which is $O(|Y||D|)$ times of computing time of myopic strategy. Therefore, it is not worthy trying 2-trial global strategy which couldn't substantially increase global expected utility but needs cumbersome computation.

## 5. Mathematical Recursion for $T$-trial Global Strategy

Previous simulations turn out that 2-trial global strategy couldn't substantially and fundamentally improve the MSE of estimated parameters and information gain, because the expected utility of the 2$^{nd}$ step ahead contributes marginally compared to the 1$^{st}$ step ahead in the optimal global utility. According to Eq.(5), 3-trial global strategy would be rewritten recursively by backward dynamic programming as Eq.(8-10).

For each possible $d_t \in D, y_t \in Y, d_{t+1} \in D$ and $y_{t+1} \in Y$,

$$V_{t+2}\left(p_{t+2}(\Theta|d_{t+1}, y_{t+1}, d_t, y_t)\right) = \max_{d_{t+2} \in D}\{u\left(d_{t+2}|p_{t+2}(\Theta|d_{t+1}, y_{t+1}d_t, y_t)\right)\}, \tag{8}$$

For each possible $d_t \in D, y_t \in Y$,
$$V_{t+1}\left(p_{t+1}(\Theta|d_t, y_t)\right) = \max_{d_{t+1} \in D}\left(u\left(d_{t+1}\Big|p_{t+1}(\Theta|d_t, y_t)\right) +\right.$$
$$\left.\sum_{y_{t+1}} p(y_{t+1}|p_{t+1}(\Theta|d_t, y_t), d_{t+1}) * V_{t+2}\left(p_{t+2}(\Theta|d_{t+1}, y_{t+1}d_t, y_t)\right)\right), \tag{9}$$
$$V_t\left(p_t(\Theta)\right) = \max_{d_t \in D}\left(u\left(d_t\Big|p_t(\Theta)\right) + \sum_{y_t} p(y_t|p_t(\Theta), d_t) * V_{t+1}\left(p_{t+1}(\Theta|d_t, y_t)\right)\right). \tag{10}$$

Here $p_{t+2}(\Theta|d_{t+1}, y_{t+1}d_t, y_t)$ is the prior distribution updated from $d_t, y_t, d_{t+1}$ and $y_{t+1}$. Given $p_{t+1}(\Theta|d_t, y_t)$, Eq.(8-9) are 2-trial global strategy and we assume that the difference between $V_{t+1}\left(p_{t+1}(\Theta|d_t, y_t)\right)$ and $\max_{d_{t+1} \in D}\left(u\left(d_{t+1}\Big|p_{t+1}(\Theta|d_t, y_t)\right)\right)$ is minor according to simulations in the previous sections. Eq.(9-10) can be treated as 2-trial global strategy too from the aspect of recursion, but expectation computation term $\sum_{y_t} p_r(y_t|p_t(\Theta), d_t) * V_{t+1}\left(p_{t+1}(\Theta|d_t, y_t)\right)$ in Eq.(10) smoothen the 'maximal value' of $V_{t+1}\left(p_{t+1}(\Theta|d_t, y_t)\right)$ in Eq.(9). The contribution of $V_{t+2}(p_{t+2}(\Theta|d_{t+1}, y_{t+1}d_t, y_t))$ to global utility improvement is getting even more negligible. Therefore, by backward recursion of Eq.(10) with the maximization in Eq.(9), we can infer that utility improvement from 2-trial global strategy to 3-trial global strategy should be less than from myopic strategy to 2-trial global strategy. Moreover, it can be inferred that $\max_{d \in D} u(d|p_t(\Theta))$ dominates the value of global expected utility for 3-trial global strategy.

In the same way, we can recursively infer that utility improvement gained from $T$-1-trial global strategy to $T$-trial global strategy ($T \geq 3$) is more marginal than from $T$-2-trial global strategy to $T$-1-trial global strategy. Similarly, the currently maximal myopic $t$-th trial utility, *i.e.*, $\max_{d \in D} u(d|p_t(\Theta))$, dominates the optimal global expected utility for $T$-trial global strategy.

By mathematical reduction, it can be generalized that the contribution of the utility improvement from the future steps keep decreasing as the horizon moves further. On the other hand, we believe the global expected utility wouldn't increase if we handle the global optimization strategy with approximation, since the approximation of posterior $p_t(\Theta)$ may negatively affect the objective of global strategy.

## 6. Conclusion

This study has shown that multiple-trail global strategy should not be expected to substantially outperform myopic one-step ahead strategy for Bayesian parameter estimation based on experimental simulations and mathematical recursion. This is because the contribution to the global information gain from those horizon steps beyond the immediate one is marginal compared to the information gain from the immediate step, and more specifically, the further into the future, the smaller contribution that future horizon step would bring. Considering that, researchers need to decide whether the amount of efforts invested to achieve global optimization is worthwhile to gain what may be a negligible lift compared with the computation efficiency.

## References

Bertsekas, D. P. (2012). *Dynamic programming and optimal control, vol. II: Approximate dynamic.* MA: Athena Scienti.

Cashore, J. M., Kumarga, L., & Frazier, P. I. (2016). Multi-step Bayesian optimization for one-dimensional feasibility determination. Retrieved from https://arxiv.org/abs/1607.03195.

Cavagnaro, D., Myung, J., Pitt, M., Kujala, J. (2010). Adaptive design optimization: a mutual information based approach to model discrimination in cognitive science. *Neural Computation*, 22, 887-905. doi: 10.1162/neco.2009.02-09-959

Garnett, R., Krishnamurthy, Y., & Wang, Donghan.( 2011). Bayesian optimal active search on Graphs. *Proceedings of the Ninth Workshop on Mining and Learning with Graphs*.

Garnett, R., Krishnamurthy, Y., & Xiong, X. (2012). Bayesian optimal active search and surveying. *Proceedings of the 29-th International Conference on Machine Learning.* Edinburgh, Scotland, UK.

Jiang, S., Chai H., Gonzalez, J., & Garnett, R. (2020). BINOCULARS for efficient nonmyopic sequential experimental design. Retrieved from https://arxiv.org/abs/1909.04568.

Kandasamy, K., Schneider, J., & Poczos, B. (2017). Query efficient posterior estimation in scientific experiments via Bayesian active learning. *Artificial Intelligence Journal*, 43, 45-56. doi: 10.1016/j.artint.2016.11.002

Kontsevich, L. L., & Tyler, C. W. (1999). Bayesian adaptive estimation of psychometric slope and threshold. *Vision Research,* 39, 2729–2737. doi: 10.1016/s0042-6989(98)00285-5

Kujala, J., & Lukka, T. J. (2006). Bayesian adaptive estimation: The next dimension. *Journal of Mathematical Psychology,* 50, 369-389. doi: 10.1016/j.jmp.2005.12.005

Kuss, M., Ja*a*kel, F., & Wichmann, F. A. (2005). Bayesian inference for psychometric functions. *Journal of Vision*, 5, 478-492. doi: 10.1167/5.5.8

Lesmes, L., Lu, Z., Baek, J., & Albright, T. (2010). Bayesian adaptive estimation of the contrast sensitivity function: The quick CSF method. *Journal of Vision*, 17, 1-21. doi: 10.1167/10.3.17

Lindley, D. V. (1956). On a measure of the information provided by an experiment. *Ann. Math. Stat.,* 27, 986-1005.

Powell, W. B. (2011). *Approximate Dynamic Programming: Solving the curses of dimensionality.* New York: John Wiley and Sons.

Truong, V. A. (2014). Approximation algorithm for the stochastic multi-period inventory problem via a look-ahead optimization approach. *Mathematics of Operations Research,* 39, 1039-1056. doi: 10.1287/moor.2013.0639

Watson, A. B. & Pelli, D. G. (1983). QUEST: a Bayesian adaptive psychometric method. *Perception & Psychophysics,* 33, 113–120. doi: 10.3758/BF03202828

Wu, J., & Frazier, P. (2019). Practical two-step lookahead Bayesian optimization. *NIPS Proceedings.*

Yue, X., & Kontar, R. A.（2020). Why non-myopic Bayesian optimization is promising and how far should we lookahead? A study via rollout. *AISTATS (Accepted).*

Zhu, J. P., & Zhang, K. L. (2015）．A Bayesian adaptive inference approach to estimating heterogeneous gap acceptance functions. *Operations and traffic management, Transportation Research Board.* Washington D.C., USA.